\documentclass[11pt,a4paper]{article}
\usepackage[hyperref]{acl2019}
\usepackage{times}
\usepackage{latexsym}

\usepackage{url}
\usepackage{booktabs}
\usepackage{amsmath}
\usepackage{amssymb}
\usepackage{multirow}
\usepackage{graphicx}
\usepackage[percent]{overpic}
\usepackage{wrapfig}
\usepackage{graphbox}

\usepackage{todonotes}
\usepackage{soul}

\allowdisplaybreaks

\aclfinalcopy 

\title{Effective Cross-lingual Transfer of Neural Machine Translation Models without Shared Vocabularies}

\author{Yunsu Kim \hspace{9pt} Yingbo Gao \hspace{9pt} Hermann Ney\\
  Human Language Technology and Pattern Recognition Group \\
  RWTH Aachen University, Aachen, Germany \\
  {\tt \{surname\}@cs.rwth-aachen.de} \\}

\date{}

\begin{document}
\maketitle
\begin{abstract}
  Transfer learning or multilingual model is essential for low-resource neural machine translation (NMT), but the applicability is limited to cognate languages by sharing their vocabularies. This paper shows effective techniques to transfer a pre-trained NMT model to a new, unrelated language without shared vocabularies. We relieve the vocabulary mismatch by using cross-lingual word embedding, train a more language-agnostic encoder by injecting artificial noises, and generate synthetic data easily from the pre-training data without back-translation. Our methods do not require restructuring the vocabulary or retraining the model. We improve plain NMT transfer by up to +5.1\% \textsc{Bleu} in five low-resource translation tasks, outperforming multilingual joint training by a large margin. We also provide extensive ablation studies on pre-trained embedding, synthetic data, vocabulary size, and parameter freezing for a better understanding of NMT transfer.\\
\end{abstract}

\section{Introduction}

Despite recent success of neural machine translation (NMT) \cite{bahdanau2014neural,vaswani2017attention}, its major improvements and optimizations cannot be easily applied to low-resource language pairs. Basic training procedure of NMT does not function well with only a handful of bilingual data \cite{koehn2017six}, while collecting bilingual resource is arduous for many languages.

Multilingual NMT solves the problem of lacking bilingual data by training a shared model along with other related languages \cite{firat2016multi,johnson2017google}. For this to work in practice, however, we need a considerable effort to gather bilingual data over multiple languages and preprocess them jointly before training. This has two critical issues: 1) The languages for training should be linguistically related in order to build a shared vocabulary. 2) It is not feasible to add a new language to a trained model, since the training vocabulary must be redefined; one may need to re-train the model from scratch.


In transfer learning \cite{zoph2016transfer}, adapting to a new language is conceptually simpler; given an NMT model pre-trained on a high-resource language pair (\emph{parent}), we can just continue the training with bilingual data of another language pair (\emph{child}). Here, the vocabulary mismatch between languages is still a problem, which seriously limits the performance especially for distant languages.

This work proposes three novel ideas to make transfer learning for NMT widely applicable to various languages:
\begin{itemize}\itemsep0em
    \item We alleviate the vocabulary mismatch between parent and child languages via cross-lingual word embedding.
    \item We train a more general encoder in the parent training by injecting artificial noises, making it easier for the child model to adapt to.
    \item We generate synthetic data from parallel data of the parent language pair, improving the low-resource transfer where the conventional back-translation \cite{sennrich2016improving} fails.
\end{itemize}
These techniques give incremental improvements while we keep the transfer unsupervised, i.e. it does not require bilingual information between the transferor and the transferee. Note that adapting to a new language is done without shared vocabularies; we need neither to rearrange joint subword units nor to restart the parent model training.

Experiments show that our methods offer significant gain in translation performance up to +5.1\% \textsc{Bleu} over plain transfer learning, even when transferring to an unrelated, low-resource language. The results significantly outperform multilingual joint training \cite{johnson2017google} in all of our experiments. We also provide in-depth analyses of the following aspects to understand the behavior of NMT transfer and maximize its performance: type of the pre-trained embedding, synthetic data generation methods, size of the transferred vocabulary, and parameter freezing.

\section{Neural Machine Translation}

Before describing our transfer learning approach, this section covers basics of an NMT model. Explanations here are not based on a specific architecture but extendable to more complex model variants.

For a source sentence $f_1^J=f_1,...,f_j,...,f_J$ (length $J$) and a corresponding target sentence $e_1^I=e_1,...,e_i,...,e_I$ (length $I$), NMT models the probability $p(e_1^I|f_1^J)$ with several components: source/target word embeddings, an encoder, a decoder, and an output layer.

Source word embedding $E^\mathrm{src}$ maps a discrete word $f$ (as a one-hot vector) to a continuous representation (\emph{embedding}) of that word $E^\mathrm{src}(f)$. In practice, it is implemented by a lookup table and stored in a matrix in $\mathbb{R}^{D \times V^\mathrm{src}}$, where $D$ is the number of dimensions of the embedding. Target word embedding is analogous.

An encoder takes a sequence of source word embeddings $E^\mathrm{src}(f_1^J)$ and produces a sequence of hidden representations $\mathbf{h}_1^J$ for the source sentence. The encoder can be modeled with recurrent \cite{sutskever2014sequence}, convolutional \cite{gehring2017convolutional}, or self-attentive layers \cite{vaswani2017attention}. The encoder is responsible for modeling syntactic and semantic relationships among the source words, including word order. 

A decoder generates target words for each target position $i$ from its internal state $\mathbf{s}_i$, which depends on $\mathbf{h}_1^J$, $E^\mathrm{tgt}(e_{i-1})$, and $\mathbf{s}_{i-1}$. It keeps track of the generated hypothesis up to position $i$-1 and relates the generation with source representations $\mathbf{h}_1^J$. For shared vocabularies between source and target languages, the target embedding weights can be tied with the source embedding weights, i.e. $E^\mathrm{src}=E^\mathrm{tgt}$.

The model is trained on a parallel corpus by optimizing for the cross-entropy loss with the stochastic gradient descent algorithm. Translation is carried out with a beam search. For more details, we refer the reader to \newcite{bahdanau2014neural} and \newcite{vaswani2017attention}.

\section{Transfer Learning for NMT}

In general, transfer learning is reusing the knowledge from other domains/tasks when facing a new problem \cite{thrun2012learning}. It has been of continued interest in machine learning for the past decades, especially when there is not enough training data for the problem at hand. Much attention is given to transfer learning for neural networks, since hidden layers of the network can implicitly learn general representations of data; the knowledge can be readily transferred by copying the hidden layer weights to another network \cite{caruana1995learning,bengio2012deep}.

For NMT, the easiest case of transfer learning is across text domains. Having an NMT model trained on some data, we can continue the training from the same network parameters with data from another domain \cite{luong2015stanford,freitag2016fast}. Transfer from another natural language processing task is also straightforward; for example, we can initialize the parameters of NMT models with pre-trained language models of corresponding languages, since the encoder and decoder are essentially language models except a few additional translation-specific components \cite{ramachandran2017unsupervised,lample2019cross}.

\begin{figure}[!ht]
    \centering
    \includegraphics[width=0.9\linewidth]{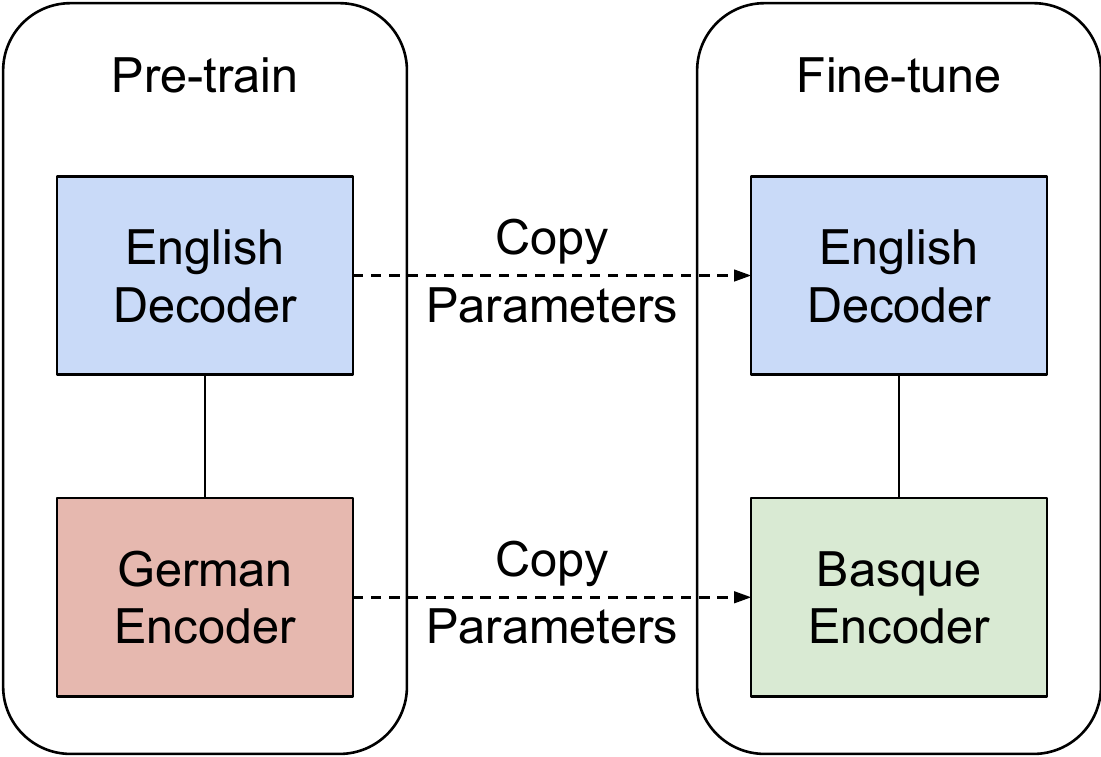}
    \caption{Diagram of transfer learning for NMT from German$\rightarrow$English to Basque$\rightarrow$English.}
    \label{fig:transfer}
\end{figure}

However, it is inherently difficult to transfer NMT models between languages, i.e. pre-train a model for a high-resource language pair and use the trained parameters for a low-resource language pair (Figure \ref{fig:transfer}). Changing a language introduces a completely different data space that does not fit to the pre-trained model. In the following, we describe this discrepancy in detail and propose our solutions. We focus on switching source languages, while the target language is fixed.

\subsection{Cross-lingual Word Embedding}\label{sec:cross}

The biggest challenge of cross-lingual transfer is the vocabulary mismatch. A natural language vocabulary is discrete and unique for each language, while the mapping between two different vocabularies is non-deterministic and arbitrary. Therefore, when we merely replace a source language, the NMT encoder will see totally different input sequences; pre-trained encoder weights do not get along with the source embedding anymore.

A popular solution to this is sharing the vocabulary among the languages of concern \cite{nguyen2017transfer,kocmi2018trivial}. This is often implemented with joint learning of subword units \cite{sennrich2016neural}. Despite its effectiveness, it has an intrinsic problem in practice: A parent model must be trained already with a shared vocabulary with child languages. Such a pre-trained parent model can be transferred only to those child languages using the same shared vocabulary. When we adapt to a new language whose words are not included in the shared vocabulary, we should learn a joint subword space again with the new language and retrain the parent model accordingly---very inefficient and not scalable.

A shared vocabulary is also problematic in that it must be divided into language-specific portions. When many languages share it, an allocated portion for each will be smaller and accordingly less expressive. This is the reason why the vocabulary is usually shared only for linguistically related languages, effectively increasing the portion of common surface forms.

In this work, we propose to keep the vocabularies separate, but share their embedding spaces instead of surface forms. This can be done independently from the parent model training and requires only monolingual data of the child language:
\begin{enumerate}\itemsep0em
    \item Learn monolingual embedding of the child language $E^\mathrm{mono}_\mathrm{child}$, using e.g. the skip-gram algorithm \cite{mikolov2013efficient}.
    \item Extract source embedding $E^\mathrm{src}_\mathrm{parent}$ from a pre-trained parent NMT model.
    \item Learn a cross-lingual linear mapping $W\in\mathbb{R}^{D\times D}$ between 1 and 2 by minimizing the objective below:
    \begin{align}
        \sum_{(f,f')\in S}\|W E^\mathrm{mono}_\mathrm{child}(f) - E^\mathrm{src}_\mathrm{parent}(f')\|_2
    \end{align}
    \item Replace source embedding of the parent model parameters with the learned cross-lingual embedding.
    \begin{align}
    E^\mathrm{src}_\mathrm{parent} \leftarrow W E^\mathrm{mono}_\mathrm{child}
    \end{align}
    \item Initialize the child model with 4 and start the NMT training on the child language pair.
\end{enumerate}

\begin{figure}[!t]
    \centering
    \begin{overpic}[width=0.9\linewidth]{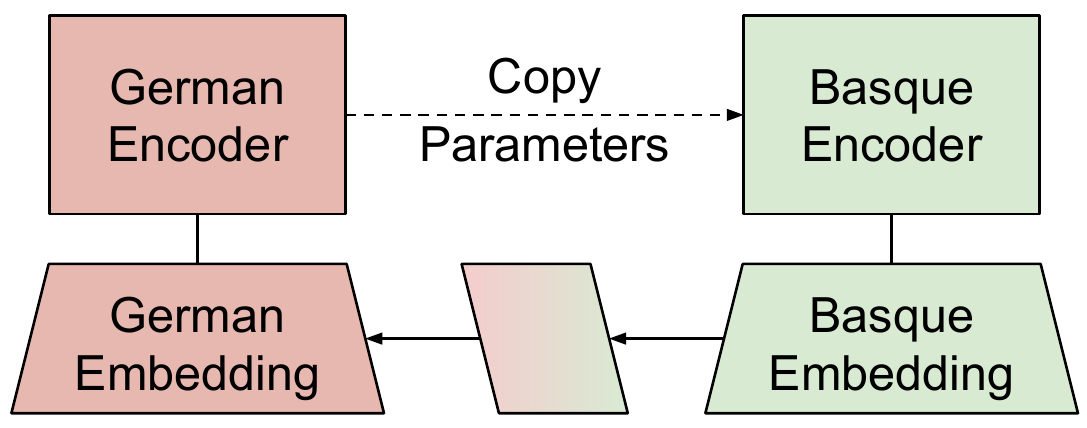}
    \put (47,7) {$W$}
    \end{overpic}
    \caption{Cross-lingual mapping of a child (Basque) embedding to the parent (German) embedding.}
    \label{fig:cross-embed}
\end{figure}

The dictionary $S$ in Step 3 can be obtained in an unsupervised way by adversarial training \cite{conneau2018word} or matching digits between the parent and child languages \cite{artetxe2017learning}. The mapping $W$ can be also iteratively refined with self-induced dictionaries of mutual parent-child nearest neighbors \cite{artetxe2017learning}, which is still unsupervised. The cross-lingually mapped child embeddings fit better as input to the parent encoder, since they are adjusted to a space similar to that of the parent input embeddings (Figure \ref{fig:cross-embed}).

Note that in Step 4, the mapping $W$ is not explicitly inserted as additional parameters in the network.
It is multiplied by $E^\mathrm{mono}_\mathrm{child}$ and the result is used as the initial source embedding weights.
The initialized source embedding is also fine-tuned along with the other parameters in the last step.

These steps do not involve rearranging a joint vocabulary or retraining of the parent model. Using our method, one can pre-train a single parent model once and transfer it to many different child languages efficiently.

Our method is also effective for non-related languages that do not share surface forms, since we address the vocabulary mismatch in the embedding level. After each word is converted to its embedding, it is just a continuous-valued vector in a mathematical space; matching vocabularies is done by transforming the vectors irrespective of language-specific alphabets.

\begin{figure}[!ht]
    \centering
    \includegraphics[width=0.9\linewidth]{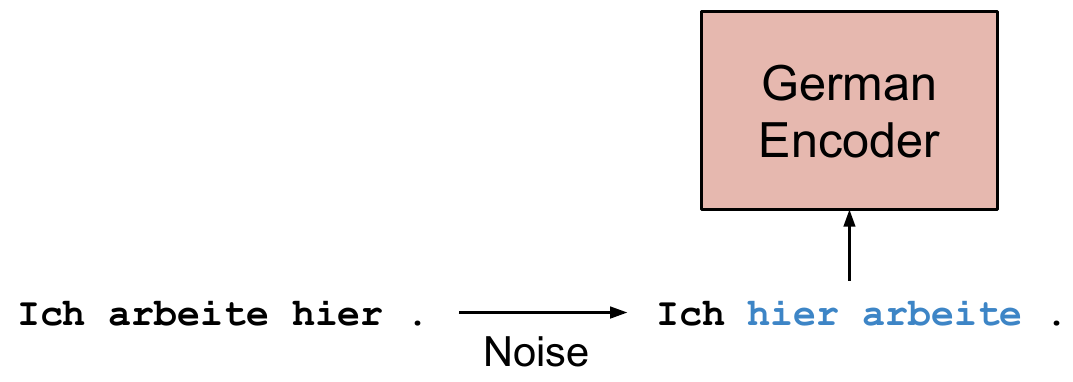}
    \caption{Injecting noise into a German (parent) source sentence.}
    \label{fig:noise}
\end{figure}

\subsection{Artificial Noises}

Another main difference between languages is the word order, namely syntactic structure of sentences. Neural sequence-to-sequence models are highly dependent on sequential ordering of the input, i.e. absolute/relative positions of input tokens.

When we train an encoder for a language, it learns the language-specific word order conventions, e.g. position of a verb in a clause, structure of an adverb phrase, etc. If the input language is changed, the encoder should adjust itself to unfamiliar word orders. The adaptation gets more difficult for non-related languages.

To mitigate this syntactic difference in cross-lingual transfer for NMT, we suggest to generalize the parent encoder so that it is not overoptimized to the parent source language. We achieve this by modifying the source side of the parent training data, artificially changing its word orders with random noises (Figure \ref{fig:noise}). The noise function includes \cite{hill2016learning,kim2018improving}:
\begin{itemize}\itemsep0em
    \item Inserting a word between original words uniformly with a probability $p_\mathrm{ins}$ at each position, choosing the inserted word uniformly from the top $V_\mathrm{ins}$ frequent words
    \item Deleting original words uniformly with a probability $p_\mathrm{del}$ at each position
    \item Permuting original word positions uniformly within a limited distance $d_\mathrm{per}$
\end{itemize}

The noises are injected into every source sentence differently for each epoch. The encoder then sees not only word orders of the parent source language but also other various sentence structures. Since we set limits to the randomness of the noises, the encoder is still able to learn general monotonicity of natural language sentences. This makes it easier for the parent encoder to adapt to a child source language, effectively transferring the pre-trained language-agnostic knowledge of input sequence modeling.

\subsection{Synthetic Data from Parent Model Training Data}\label{sec:syn}

Transfer learning for NMT is particularly necessary for low-resource language pairs where the bilingual data is scarce. The standard technique to address the scarcity is generating synthetic parallel data from target monolingual corpora via back-translation \cite{sennrich2016improving}. However, this works only if the generated source sentences are of sufficiently acceptable quality. In low-resource translation tasks, it is hard to train a good target-to-source translation model, which is used to produce the source hypotheses.

\begin{figure}[!ht]
    \centering
    \includegraphics[width=0.8\linewidth]{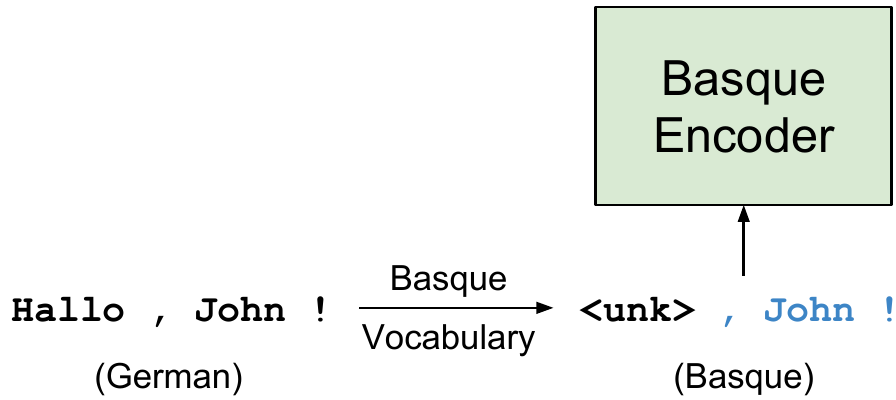}
    \caption{Synthetic Basque sentence generated from a German sentence.}
    \label{fig:synthetic}
\end{figure}

For these scenarios, we devise a simple trick to create additional parallel data for the child language pair without training a target-to-source translation model. The idea is to reuse the parallel data already used for training the parent model. In the source side, we retain only those tokens that exist in the child vocabulary and replace all other tokens with a predefined token, e.g. \texttt{<unk>} (Figure \ref{fig:synthetic}). The target side stays the same as we do not switch the languages.

The source side of this synthetic data consists only of the overlapping vocabulary entries between the parent and child languages. By including this data in the child model training, we prevent an abrupt change of the input to the pre-trained model while keeping the parent and child vocabularies separated. It also helps to avoid overfitting to a tiny parallel data of the child language pair.

In addition, we can expect a synergy with cross-lingual word embedding (Section \ref{sec:cross}), where the source embedding space of the child task is transformed into that of the parent task. In this cross-lingual space, an overlapping token between parent and child vocabularies should have a very similar embedding to that in the original parent embedding space, to which the pre-trained encoder is already familiar. This helps to realize a smooth transition from parent source input to child source input in the transfer process.

\section{Main Results}
\label{sec:main}

We verify the effect of our techniques in transfer learning setups with five different child source languages: Basque (eu), Slovenian (sl), Belarusian (be), Azerbaijani (az), and Turkish (tr). Target language is fixed to English (en) and we use German$\rightarrow$English as the parent language pair.\vspace{0.7em}

\noindent\textbf{Data}: The parent model was trained on parallel data of WMT 2018 news translation task\footnote{http://www.statmt.org/wmt18/translation-task.html} and synthetic data released by \newcite{sennrich2016edinburgh}. For the child language pairs, we used IWSLT 2018 low-resource MT task data (eu-en) \cite{jan2018iwslt}, IWSLT 2014 MT task data (sl-en) \cite{cettolo2014report}, TED talk data from \cite{qi2018and} (be-en/az-en), and subsampling of WMT 2018 news translation task data (tr-en). Statistics of the parallel corpora are given in Table \ref{tab:stat}. Note that the child source languages are linguistically far from the parent source.
\begin{table}[!t]
    \centering
    \begin{tabular}{ccr}
        \toprule
        & Source & \multicolumn{1}{c}{Data ($\rightarrow$English)}\\
        Family & Language & \multicolumn{1}{c}{[\#sents]}\\
        \midrule
        Germanic & German & 10,111,758\enspace\enspace\\
        \midrule
        Isolate & Basque & 5,605\enspace\enspace\\
        \midrule
        \multirow{2}{*}{Slavic} & Slovenian & 17,103\enspace\enspace\\
        & Belarusian & 4,509\enspace\enspace\\
        \midrule
        \multirow{2}{*}{Turkic} & Azerbaijani & 5,946\enspace\enspace\\
        & Turkish & 9,998\enspace\enspace\\
        \bottomrule
    \end{tabular}
    \caption{Language families and parallel data statistics.}
    \label{tab:stat}
\end{table}

Every training dataset was preprocessed with the Moses tokenizer\footnote{http://www.statmt.org/moses/}, where the source side was lowercased and the target side was frequent-cased.\vspace{0.7em}

\begin{table*}[!t]
    \centering
    \begin{tabular}{lcccccc}
        \toprule
        & \multicolumn{5}{c}{\textsc{Bleu} [\%]}\\
        System & eu-en & sl-en & be-en & az-en & tr-en \\
        \midrule
        Baseline & 1.7 & 10.1 & 3.2 & 3.1 & 0.8\\
        Multilingual \cite{johnson2017google} & 5.1 & 16.7 & 4.2 & 4.5 & 8.7\\
        \midrule
        Transfer \cite{zoph2016transfer} & 4.9 & 19.2 & 8.9 & 5.3 & 7.4\\
        + Cross-lingual word embedding & 7.4 & 20.6 & 12.2 & 7.4 & 9.4 \\
        \hspace{7pt} + Artificial noises & 8.2 & 21.3 & 12.8 & 8.1 & 10.1\\
        \hspace{17pt} + Synthetic data & \textbf{9.7} & \textbf{22.1} & \textbf{14.0} & \textbf{9.0} & \textbf{11.3} \\
        \bottomrule
    \end{tabular}
    \caption{Translation results of different transfer learning setups.}
    \label{tab:results}
\end{table*}

\noindent\textbf{Transfer learning}: All NMT models in our experiments follow the base 6-layer Transformer architecture of \newcite{vaswani2017attention}, except that the source and target embedding weights are not tied. Each source language was encoded with byte pair encoding (BPE) \cite{sennrich2016neural} with 20k merge operations, while the target language was encoded with 50k BPE merges. Dropout with probability of 0.3 was applied to Transformer prepost/activation/attention components in both parent and child model trainings. Training was carried out with Sockeye \cite{hieber2017sockeye} using the Adam optimizer \cite{kingma2014adam} with the default parameters. The maximum sentence length was set to 100 and the batch size to 4,096 words. We stopped the training when perplexity on a validation set was not improving for 12 checkpoints. We set checkpoint frequency to 10,000 updates for the parent model and 1,000 updates for the child models. The parent model yields 39.2\% \textsc{Bleu} on WMT German$\rightarrow$English newstest2016 test set.\vspace{0.7em}

\noindent\textbf{Baseline}: As a baseline child model without transfer learning, we used the same setting as above but learned a shared source-target BPE vocabulary with 20k merge operations. We also tied source and target embeddings as suggested for low-resource settings in \newcite{schamper2018rwth}. Dropout was applied also to the embedding weights for the baselines.\vspace{0.7em}

\noindent\textbf{Multilingual}: We also compare our transfer learning with the multilingual training where a single, shared NMT model is trained for the parent and child language pairs together from scratch \cite{johnson2017google}. For each child task, we learned a joint BPE vocabulary of all source and target languages in the parent/child tasks with 32k merge operations. The training data for the child task was oversampled so that each mini-batch has roughly 1:1 ratio of the parent/child training examples.

Note that we built a different multilingual model for each child task. Since they depend on shared vocabularies, we should restructure the vocabulary and retrain the model for each of the new language pairs we wish to adapt to.
\vspace{0.7em}

\noindent\textbf{Cross-lingual word embedding}: To pre-train word embeddings, we used Wikimedia dumps\footnote{https://dumps.wikimedia.org/} of timestamp 2018-11-01 for all child languages except Turkish for which we used WMT News Crawl 2016-2017. From Wikimedia dumps, the actual articles were extracted first\footnote{https://github.com/attardi/wikiextractor/}, which were split to sentences using the StanfordCoreNLP toolkit \cite{manning2014stanford}. Monolingual embeddings were trained with fasttext \cite{bojanowski2017enriching} with minimum word count 0. For learning the cross-lingual mappings, we ran 10 epochs of adversarial training and another 10 epochs of dictionary-based refinement using MUSE \cite{conneau2018word}. We chose top 20k types as discriminator inputs and 10k as maximum dictionary rank.\vspace{0.7em}

\noindent\textbf{Artificial noises}: Following \newcite{kim2018improving}, we used these values for the noise model: $p_\mathrm{ins} = 0.1$, $V_{ins} = 50$, $p_\mathrm{del} = 0.1$, and $d_\mathrm{per} = 3$. We empirically found that these values are optimal also for our purpose. The parent model trained with noises gives 38.2\% \textsc{Bleu} in WMT German$\rightarrow$English newstest2016: 1.0\% worse than without noises.\vspace{0.7em}

\noindent\textbf{Synthetic data}: We uniformly sampled 1M sentence pairs from German$\rightarrow$English parallel data used for the parent training and processed them according to Section \ref{sec:syn}. The child model parallel data was oversampled to 500k sentence pairs, making an overall ratio of 1:2 between the parallel and synthetic data. We also tried other ratio values, e.g. 1:1, 1:4, or 2:1, but the performance was consistently worse.\\

Table \ref{tab:results} presents the results. Plain transfer learning already gives a boost but is still far from a satisfying quality, especially for Basque$\rightarrow$-English and Azerbaijani$\rightarrow$English. On top of that, each of our three techniques offers clear, incremental improvements in all child language pairs with a maximum of 5.1\% \textsc{Bleu} in total.

Cross-lingual word embedding shows a huge improvement up to +3.3\% \textsc{Bleu}, which exhibits the strength of connecting parent-child vocabularies on the embedding level. If we train the parent model with artificial noises on the source side, the performance is consistently increased by up to +0.8\% \textsc{Bleu}. This occurs even when dropout is used in the parent model training; randomizing word orders provides meaningful regularization which cannot be achieved via dropout. Finally, our synthetic data extracted from the parent parallel data is proved to be effective in low-resource transfer to substantially different languages: We obtain an additional gain of at most +1.5\% \textsc{Bleu}.

Our results also surpass the multilingual joint training by a large margin in all tasks. One shared model for multiple language pairs inherently limits the modeling capacity for each task. Particularly, if one language pair has much smaller training data than the other, oversampling the low-resource portion is not enough to compensate the scale discrepancy in multilingual training. Transfer learning with our add-on techniques is more efficient to exploit knowledge of high-resource language pairs and fine-tune the performance towards a child task.

\section{Analysis}

In this section, we further investigate our methods in detail in comparison to their similar variants, and also perform ablation studies for the NMT transfer in general.

\subsection{Types of Pre-trained Embedding}

\begin{table}[!h]
    \centering
    \begin{tabular}{lcc}
        \toprule
        Pre-trained embedding & \textsc{Bleu} [\%]\\
        \midrule
        None & 5.3\\
        Monolingual & 6.3\\
        Cross-lingual (az-de) & \textbf{7.4}\\
        Cross-lingual (az-en) & 7.1\\
        \bottomrule
    \end{tabular}
    \caption{Azerbaijani$\rightarrow$English translation results with different types of pre-trained source embeddings.}
    \label{tab:embed}
\end{table}
We analyze the effect of the cross-linguality of pre-trained embeddings in Table \ref{tab:embed}. We observe that monolingual embedding without a cross-lingual mapping also improves the transfer learning, but is significantly worse than our proposed embedding, i.e. mapped to the parent source (de) embedding. The mapping can be learned also with the target (en) side with the same procedure as in Section \ref{sec:cross}. The target-mapped embedding is not compatible with the pre-trained encoder but directly guides the child model to establish the connection between the new source and the target. It also improves the system, but our method is still the best among the three embedding types.

\subsection{Synthetic Data Generation}

\begin{table}[!h]
    \centering
    \begin{tabular}{lccc}
        \toprule
        Synthetic data & \textsc{Bleu} [\%]\\
        \midrule
        None & 8.2\\
        Back-translation & 8.3\\
        Empty source & 8.2 \\
        Copied target & 8.9\\
        Parent model data & \textbf{9.7}\\
        + Cross-lingual replacement & 8.7\\
        \bottomrule
    \end{tabular}
    \caption{Basque$\rightarrow$English translation results with synthetic data generated using different methods.}
    \label{tab:syn}
\end{table}
In Table \ref{tab:syn}, we compare our technique in Section \ref{sec:syn} with other methods of generating synthetic data. For a fair comparison, we used the same target side corpus (1M sentences) for all these methods.

As explained in Section \ref{sec:syn}, back-translation \cite{sennrich2016improving} is not beneficial here because the generated source is of too low quality. Empty source sentence is proposed along with back-translation as its simplification, which does not help either in transfer learning. Copying target sentences to the source side is yet another easy way to obtain synthetic data \cite{currey2017copied}. It gives an improvement to a certain extent; however, our method of using the parent model data works much better in transfer learning.

We manually looked at the survived tokens in the source side of our synthetic data. We observed lots of overlapping tokens over the parent and child source vocabularies even if they were not shared: 4,487 vocabulary entries between Basque and German. Approximately 2\% of them are punctuation symbols and special tokens, 7\% are digits, and 62\% are made of Latin alphabets, a large portion of which is devoted to English words (e.g. named entities) or their parts. The rest of the vocabulary is mostly of noisy tokens with exotic alphabets.

As Figure \ref{fig:synthetic} illustrates, just punctuation symbols and named entities can already define a basic structure of the original source sentence. Such tokens play the role of anchors in translation; they are sure to be copied to the target side. The surrounding \texttt{<unk>} tokens are spread according to the source language structure, whereas merely copying the target sentence to the source \cite{currey2017copied} ignores the structural difference between source and target sentences. Note that our trick applies also to the languages with completely different alphabets, e.g. Belarusian and German (see Table \ref{tab:results}).

We also tested an additional processing for our synthetic data to reduce the number of unknown tokens. We replaced non-overlapping tokens in the German source side with the closest Basque token in the cross-lingual word embedding space. The result is, however, worse than not replacing them; we noticed that this subword-by-subword translation produces many Basque phrases with wrong BPE merges \cite{kim2018improving}.

\subsection{Vocabulary Size}
\begin{table}[!h]
    \centering
    \begin{tabular}{lccc}
        \toprule
        & \multicolumn{2}{c}{\textsc{Bleu} [\%]}\\
        BPE merges & sl-en & be-en\\
        \midrule
        10k & \textbf{21.0} & 11.2\\
        20k & 20.6 & \textbf{12.2}\\
        50k & 20.2 & 10.9\\
        70k & 20.0 & 10.9\\
        \bottomrule
    \end{tabular}
    \caption{Translation results with different sizes of the source vocabulary.}
    \label{tab:vocab}
\end{table}

Table \ref{tab:vocab} estimates how large the vocabulary should be for the language-switching side in NMT transfer. We varied the number of BPE merges on the source side, fixing the target vocabulary to 50k merges. The best results are with 10k or 20k of BPE merges, which shows that the source vocabulary should be reasonably small to maximize the transfer performance. Less BPE merges lead to more language-independent tokens; it is easier for the cross-lingual embedding to find the overlaps in the shared semantic space.

If the vocabulary is excessively small, we might lose too much language-specific details that are necessary for the translation process. This is shown in the 10k merges of Belarusian$\rightarrow$English.

\subsection{Freezing Parameters}
\begin{table}[!h]
    \centering
    \begin{tabular}{lccc}
        \toprule
        Frozen parameters & \textsc{Bleu} [\%]\\
        \midrule
        None & 21.0 \\
        \midrule
        Target embedding & 21.4 \\
        + Target self-attention & \textbf{22.1} \\
        \hspace{7pt} + Encoder-decoder attention & 21.8 \\
        \hspace{18pt} + Feedforward sublayer & 21.3 \\
        \hspace{30pt} + Output layer & 21.9 \\
        \bottomrule
    \end{tabular}
    \caption{Slovenian$\rightarrow$English translation results with freezing different components of the decoder.}
    \label{tab:freeze}
\end{table}

Lastly, we conducted an ablation study of freezing parent model parameters in the child training process (Table \ref{tab:freeze}). We show only the results when freezing the decoder; in our experiments, freezing any component of the encoder always degrades the translation performance. The experiments were done at the final stage with all of our three proposed methods applied.

Target embedding and target self-attention parts are independent of the source information, so it makes sense to freeze those parameters even when the source language is changed. On the contrary, encoder-decoder attention represents the relation between source and target sentences, so it should be redefined for a new source language. The performance deteriorates when freezing feedforward sublayers, since it is directly influenced by the encoder-decoder attention layer. The last row means that we freeze all parameters of the decoder; it is actually better than freezing all but the output layer.

\section{Related Work}

Transfer learning is first introduced for NMT in \newcite{zoph2016transfer}, yet with a small RNN architecture and on top frequent words instead of using subword units. \newcite{nguyen2017transfer} and \newcite{kocmi2018trivial} use shared vocabularies of BPE tokens to improve the transfer learning, but this requires retraining of the parent model whenever we transfer to a new child language.

Multilingual NMT trains a single model with parallel data of various translation directions jointly from scratch \cite{dong2015multi,johnson2017google,firat2016multi,gu2018universal}. Their methods also rely on shared subword vocabularies so it is hard for their model to adapt to a new language.

Cross-lingual word embedding is studied for the usages in MT as follows. In phrase-based SMT, \newcite{alkhouli2014vector} builds translation models with word/phrase embeddings. \newcite{kim2018improving} uses cross-lingual word embedding as a basic translation model for unsupervised MT and attach other components on top of it. \newcite{artetxe2018unsupervised} and \newcite{lample2018unsupervised} initialize their unsupervised NMT models with pre-trained cross-lingual word embeddings. \newcite{qi2018and} do the same initialization for supervised cases, observing only improvements in multilingual setups.

Artificial noises for the source sentences are used to counteract word-by-word training data in unsupervised MT \cite{artetxe2018unsupervised,lample2018unsupervised,kim2018improving}, but in this work, they are used to regularize the NMT.

\newcite{neubig2018rapid} study adapting a multilingual NMT system to a new language. They train for a child language pair with additional parallel data of its similar language pair. Our synthetic data method does not rely on the relatedness of languages but still shows a good performance. They learn just a separate subword vocabulary for the child language without a further care, which we counteract with cross-lingual word embedding.

\newcite{neubig2018param} show ablation studies on parameter sharing and freezing in one-to-many multilingual setup with shared vocabularies. Our work conduct the similar experiments in the transfer learning setting with separate vocabularies.

\newcite{platanios2018contextual} augment a multilingual model with language-specific embeddings from which the encoder and decoder parameters are inferred with additional linear transformations. They only mention its potential to transfer to an unseen language without any results on it. Our work focuses on transferring a pre-trained model to a new language without any change in the model architecture but with an explicit guidance for cross-linguality on the word embedding level.

\newcite{wang2019multilingual} address the vocabulary mismatch in multilingual NMT by using shared embeddings of character $n$-grams and common semantic concepts. Their method has a strict assumption that the languages should be related orthographically with shared alphabets, while our method is not limited to similar languages and directly benefits from advances in cross-lingual word embedding for distant languages.

Another line of research on low-resource MT is unsupervised learning \cite{lample2018unsupervised,lample2018phrase,lample2019cross,artetxe2018unsupervisedsmt,artetxe2018unsupervised,kim2018improving}, training translation models only with monolingual data. 
However, these methods are verified mostly in high-resource language pairs, e.g. French$\leftrightarrow$English, where there is no need to restrict the training data to only monolingual corpora.
In low-resource language pairs with little linguistic similarity, \newcite{neubig2018rapid} and \newcite{guzman2019two} show that unsupervised MT methods do not function at all.
We tested an unsupervised MT software \newcite{lample2019cross} internally, which also resulted in failure, e.g. 1\% \textsc{Bleu} at the Basque$\rightarrow$English task of Section \ref{sec:main}.
Moreover, unsupervised MT methods usually require a very long training time---at least 1-2 weeks with a single GPU---due to its iterative nature, while our cross-lingual transfer needs only a couple of hours of training once you have a parent model. 

Alternatively, one might consider using parallel data involving a pivot language, either by decoding in two consecutive steps \cite{kauers2002interlingua,de2006catalan,utiyama2007comparison,costa2011enhancing} or by creating pivot-based synthetic data \cite{de2006catalan,bertoldi2008phrase,zheng2017maximum,chen2017teacher}.
These methods cannot be applied to most of the language pairs from/to English, because it is extremely difficult to collect parallel data with another third language other than English.

\section{Conclusion}

In this paper, we address the problem of transferring an NMT model to unseen, unrelated language pairs. We propose three novel techniques to improve the transfer without vocabulary sharing between parent and child source languages.

Firstly, we transform monolingual embeddings of the new language into the embedding space of the parent NMT model. This accomplishes an effective transition of vocabularies on the embedding level. Secondly, we randomize the word orders in the parent model training to avoid overfitting to the parent source language. This makes it easier for the encoder to adapt to the new language syntax. For the first time, we show a practical usage of artificial noises to regularize an NMT model. Lastly, we reuse parallel data of the parent language pair in the child training phase to avoid an abrupt change of the training data distribution.

All three methods significantly improve over plain transfer learning with a total gain of up to +5.1\% \textsc{Bleu} in our experiments, consistently outperforming multilingual joint training. Our methods do not require retraining of a shared vocabulary or the parent model, enabling an incremental transfer of the same parent model to various (possibly unrelated) languages. Our implementation of the proposed methods is available online.\footnote{\scriptsize\url{https://github.com/yunsukim86/sockeye-transfer}}

As for future work, we will test our methods in the NMT transfer where the target language is switched. We also plan to compare different algorithms for learning the cross-lingual mapping \cite{artetxe2018robust,xu2018unsupervised,joulin2018loss} to optimize the transfer performance.

\section*{Acknowledgments}

\begin{center}
\vspace{0.5em}
\includegraphics[align=c,width=0.25\textwidth]{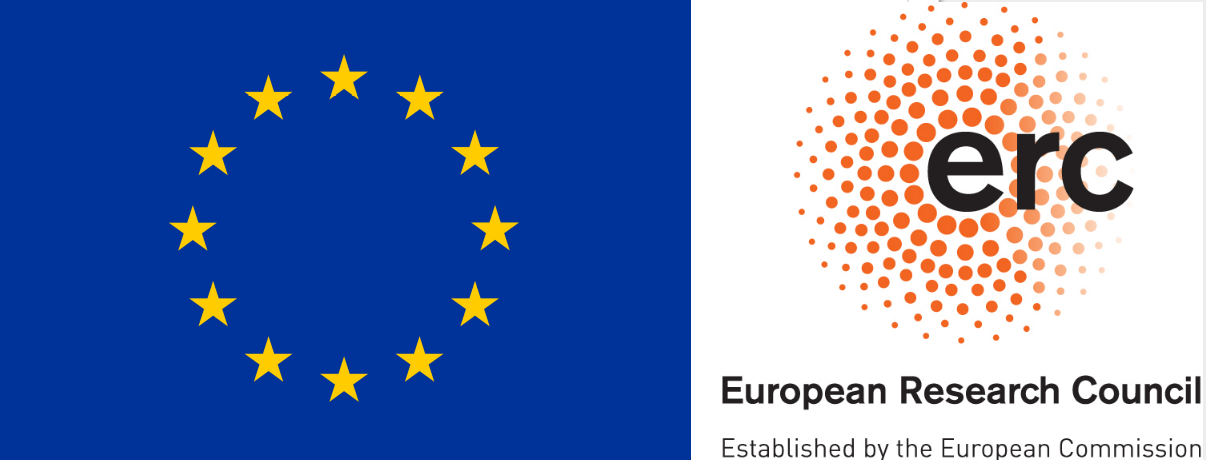}
\hspace{10pt}
\includegraphics[align=c,width=0.15\textwidth]{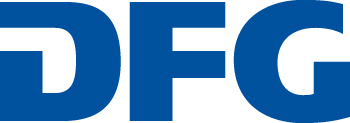}
\end{center}
\vspace{0.5em}
This work has received funding from the European Research Council (ERC) (under the European Union's Horizon 2020 research and innovation programme, grant agreement No 694537, project "SEQCLAS") and the Deutsche Forschungsgemeinschaft (DFG; grant agreement NE 572/8-1, project "CoreTec"). The GPU cluster used for the experiments was partially funded by DFG Grant INST 222/1168-1. The work reflects only the authors' views and none of the funding agencies is responsible for any use that may be made of the information it contains.

\bibliographystyle{acl_natbib}
\bibliography{references}

\end{document}